\pdfoutput=1

\documentclass[11pt]{article}

\usepackage[preprint]{acl}

\usepackage{times}
\usepackage{latexsym}
\usepackage[T1]{fontenc}
\usepackage[utf8]{inputenc}
\usepackage{graphicx}
\usepackage{inconsolata}
\usepackage{microtype}

\usepackage{multirow}
\usepackage{booktabs}
\usepackage{enumitem}
\usepackage{tcolorbox}

\newtcolorbox{promptbox}{
    colback=blue!5,  % Светло-синий фон
    colframe=blue,   % Синяя рамка
    arc=8pt,        % Закругление углов
    boxrule=1pt,    % Толщина линии
    parbox=false  % Разрешаем параграфы внутри
}

\title{PingPong: A Benchmark for Role-Playing Language Models with User Emulation and Multi-Model Evaluation}

\author{Ilya Gusev \\
  Independent researcher / Amsterdam \\
  \texttt{phoenixilya@gmail.com}
}

\begin{document}
\maketitle
\begin{abstract}
We introduce a benchmark for evaluating the role-playing capabilities of language models. Our approach leverages different language models to simulate users in dynamic, multi-turn conversations and assess the resulting dialogues. Our methodology involves three main components: a player model that adopts a specific character role, an interrogator model that simulates user behavior in a specific situation, and a judge model ensemble that evaluates conversation quality with 3 metrics: character consistency, entertainment value, and language fluency. We evaluated more than 40 models in both English and Russian, with each model participating in 64 conversations with 8 characters and 8 situations. We conducted experiments comparing automated evaluations with human annotations to validate our approach, demonstrating strong correlations across multiple criteria. This work provides a foundation for a robust and dynamic evaluation of different model capabilities in interactive scenarios.
\end{abstract}

\section{Introduction}
Language models, which predict plausible language, have dominated natural language processing since BERT~\citep{devlin-etal-2019-bert}, with models like ChatGPT~\citep{ouyang2022traininglanguagemodelsfollow} showcasing advanced conversational capabilities.

In this paper, we focus on role-playing language models for entertainment purposes. These models are assigned specific characters or personas and are tasked with maintaining these roles while engaging and entertaining users. Although there are other important applications of role-playing language models, such as training mental health specialists~\citep{health} or simulating human opinion dynamics~\citep{opinion_dyanmcis}, they are beyond the scope of this paper.

People seek entertainment through interactions with their favorite characters from games, movies, and TV shows. The quality of these interactions directly impacts user satisfaction and engagement. As companies continue to develop dedicated platforms for role-playing language models, such as Character.ai\footnote{\url{https://character.ai}} or Chai\citep{irvine2023rewardingchatbotsrealworldengagement}, robust evaluation frameworks are critical. Developers of such systems cannot effectively improve them or compare competing approaches without standardized methods to assess how well models maintain character consistency and entertain users.

\begin{figure}[t]
  \includegraphics[width=\columnwidth]{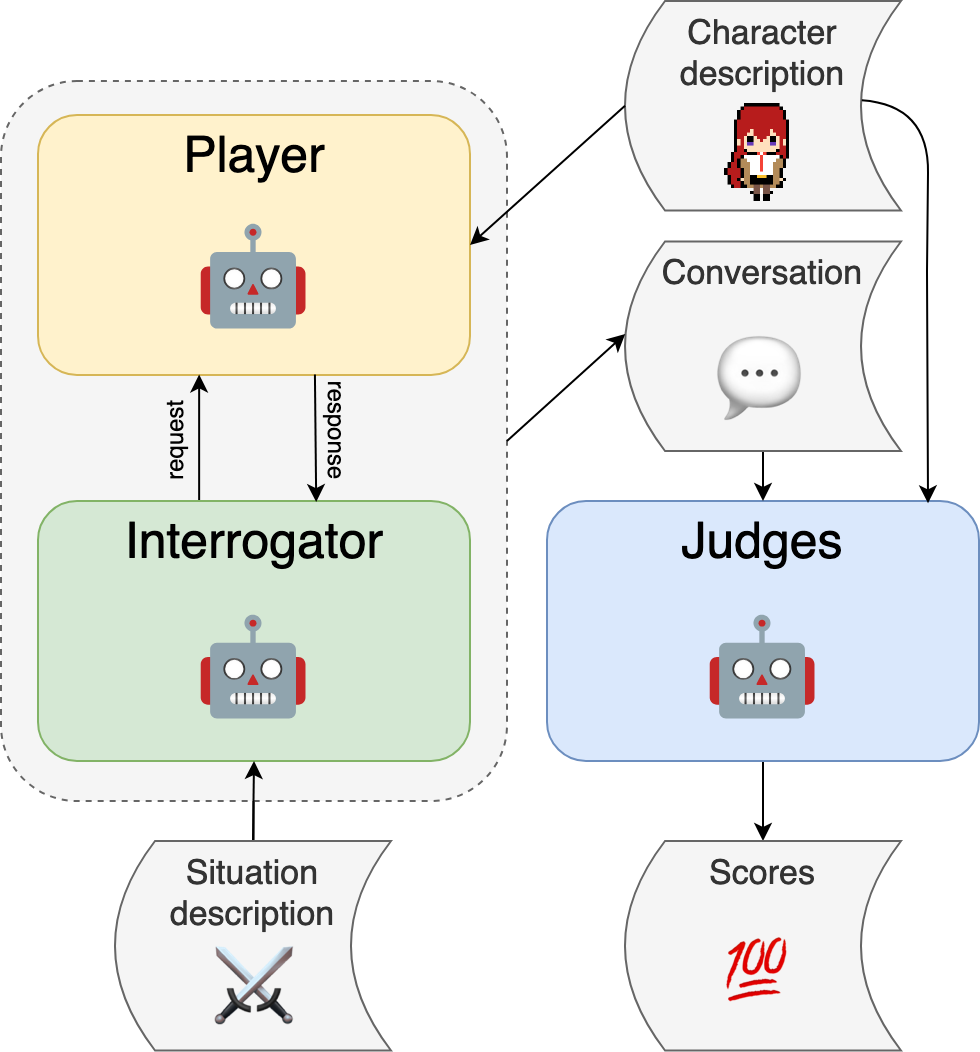}
  \caption{This diagram illustrates the flow of interactions in the proposed benchmark. There are three main components with different language models: a player, an interrogator, and a judge ensemble. The player assumes some character role, the interrogator acts as a user in a specific situation, and the judges evaluate final conversations.}
  \label{fig:overview}
\end{figure}

To fill this gap, we introduce a novel benchmark for evaluating role-playing language models. From our experience with language models, we believe that direct interaction is the most effective way to assess a language model's conversational abilities. However, humans often lack time to test new models manually, and many popular benchmarks are limited to single-turn interactions~\citep{alpacaeval2, mmlu}. These benchmarks are also becoming less reliable due to test data contamination~\citep{deng2024investigatingdatacontaminationmodern}. To address this, we propose using language models to simulate users in role-playing conversations and automatically evaluate the resulting dialogues.

Our methodology, illustrated in Figure~\ref{fig:overview}, involves three key components: a player model that assumes a character role, an interrogator model that simulates user behavior, and a judge model that evaluates the quality of the conversation. Our work builds on existing benchmarks, such as the EQ-bench~\citep{paech2023eqbench}, introducing an approach to evaluate the role-playing abilities of language models.

Our contributions:
\setlist{nolistsep}
\begin{itemize}[noitemsep]
    \item We propose a benchmark for assessing the role-playing abilities of language models. The \textbf{combination} of the following traits makes it novel: \begin{itemize}
        \item \textbf{Multi-turn}: All conversations have multiple turns to be closer to the real usage of role-playing models.
        \item \textbf{Dynamic}: The interrogator questions are generated by sampling language models and are not predefined. Each evaluation run produces different questions, making it harder for models to memorize responses and making test data contamination harder.
        \item \textbf{Multi-model}: Several judges mitigate individual model biases and correlate better with humans. It also allows for the evaluation of the models that are used as judges.
    \end{itemize}
    \item We validate our benchmark through correlation with manual annotations and comparison with other benchmarks.
    \item We show that the multi-model setup correlates better with humans than a single model.
    \item We discover that fine-tuning models for creative writing improves their role-playing abilities.
\end{itemize}

All results, prompts, and scripts are available online\footnote{\url{https://github.com/ilyagusev/ping_pong_bench}}.

The benchmark website has up-to-date leaderboards\footnote{\url{https://ilyagusev.github.io/ping_pong_bench}} and all conversations with example-wise scores, which allows the eyeballing of common errors and problems of evaluated models. It is available for English and Russian languages.

\section{Related work}

\textbf{Role-playing capabilities and evaluation.} Various commercial services exploit the role-playing abilities of language models, including Character.ai and Chai. Academic and community efforts are underway to create similar systems with open datasets, code, and models, such as PIPPA, ChatHaruhi, Character-LLM~\citep{pippa, li2023chatharuhirevivinganimecharacter, shao-etal-2023-character}, MythoMax\footnote{\url{https://huggingface.co/Gryphe/MythoMax-L2-13b}}, or Magnum\footnote{\url{https://huggingface.co/anthracite-org/magnum-v2-123b}}. Several static benchmarks for role play exist, including ECHO, InCharacter, RAIDEN, and CharacterEval~\citep{ng2024llmsechousevaluating, wang2024incharacterevaluatingpersonalityfidelity, raiden, tu2024characterevalchinesebenchmarkroleplaying}. 

PersonaGym~\citep{samuel2024personagymevaluatingpersonaagents} is close to our work, featuring dynamic question generation based on the environment (``situation'' in our terminology) and the currently selected persona. There is also a very similar dynamic benchmark, RPBench-Auto\footnote{\url{https://boson.ai/rpbench-blog}}. It is based on the same assumptions and features and has a structure similar to one of the versions of our benchmark. The significant difference from our work is that their evaluation is based on side-by-side comparisons with the baseline model responses, while we produce single-point evaluations. One can also utilize evaluation based on multiple choice questions similar to the one developed in the DITTO framework~\citep{ditto}. Although it efficiently assesses the role consistency, the evaluation is based heavily on questions based on artificial knowledge that may not reflect typical user interactions with role-playing systems.

One common problem for most of these benchmarks is a lack of long-term support and reproducibility, which might be caused by the high cost of evaluations. Regularly adding new models to a benchmark is crucial, as this is the only way the benchmark can guide practical decisions.

A different approach to evaluation would be to use online metrics, such as retention rates or user ratings~\citep{irvine2023rewardingchatbotsrealworldengagement}. However, this approach is only viable if you already have a service with a substantial user base.

\textbf{Automatic evaluation with language models.} LLM-as-a-Judge~\citep{Zheng2023JudgingLW} is an evaluation method that relies on language models, such as GPT-4, instead of humans. Popular benchmarks using this method include AlpacaEval, EQ-bench, Creative Writing, and BiGGen Bench~\citep{alpacaeval2,paech2023eqbench,kim2024biggenbenchprincipledbenchmark}. The validity of these benchmarks is based on their high correlation with human annotations, specifically with Chatbot Arena~\citep{chiang2024chatbotarenaopenplatform}.

\textbf{Multi-model evaluation.} However, all these benchmarks are based on a single model as a judge, which may introduce various biases, including self-evaluation bias~\citep{panickssery2024llmevaluatorsrecognizefavor, xu2024prideprejudicellmamplifies}. PoLL~\citep{verga2024replacingjudgesjuriesevaluating} authors aggregate evaluations from different language models similarly to us, with average pooling. They show that ensembling different models for evaluation increases the correlation with human annotations. There is another more agentic approach~\citep{chan2023chatevalbetterllmbasedevaluators} with a referee team.

\textbf{Multi-turn evaluation.} Most benchmarks are single-turn, which contrasts with the real-world usage of language models. There are multi-turn benchmarks, such as MT-Bench-101~\citep{bai2024mtbench101finegrainedbenchmarkevaluating} and MT-Eval~\citep{kwan2024mtevalmultiturncapabilitiesevaluation}, though they focus on specific capabilities, and their evaluation procedures still differ from how humans implicitly rate language models.

\textbf{Data contamination.} Another major problem for static public benchmarks is data leakage into the pre-training data sets of language models~\citep{deng2024investigatingdatacontaminationmodern}. It is challenging to avoid contamination since such tests are usually stored online and considered ``code'' during pre-training. This can occur even with no malicious intent from the model creators. The most obvious solution is to close the benchmarks completely, which requires trusting the benchmark organizers, which is difficult in a highly competitive environment. Alternative solutions include regularly updating benchmarks with new test data~\citep{white2024livebenchchallengingcontaminationfreellm} or dynamically generating test data using existing language models.

\section{Methodology}

\subsection{Role definitions}

Our framework comprises three principal roles: player, interrogator, and judge, inspired by the Turing test~\citep{alan1950a}. However, our approach differs in the number of agents, the player's objective, and the use of machine-based interrogators and judges. 

Language models can assume three possible roles.

\setlist{nolistsep}
\begin{itemize}[noitemsep]
\item \textbf{Player} assumes the role of a specific character based on a character card provided.
\item \textbf{Interrogator} engages the player within a given situation or towards a specific goal, simulating user behavior.
\item \textbf{Judge} evaluates the player's responses against predetermined criteria.
\end{itemize}

Role assignments are implemented through a combination of system and user prompts. We use only models that support chat templates. All instructions are incorporated into the user prompt for models without dedicated system prompts, such as Gemma 2~\citep{gemmateam2024gemma2improvingopen}.

This setup is \textbf{asymmetric} since the player only gets the character description, while the interrogator only gets the situation information. This is intentional, as typical use cases of role-playing models are asymmetrical. However, it is possible to modify it to make it symmetric by providing character descriptions and situations to both the player and the interrogator. Symmetrical setups might be helpful in other domains.

\subsection{Judge}
The scoring is single-point, with no reference examples or pairs. The judge used three main evaluation criteria:

\begin{itemize}
    \item \textbf{Character consistency}: The player's answers align perfectly with the assigned character; they correspond to the character's description.
    \item \textbf{Entertainment value}: The player's responses are engaging and entertaining.
    \item \textbf{Language fluency}: The language used by the player is of the highest quality and is free of errors. The player is perfectly fluent.
\end{itemize}

These criteria reflect the main things we expect from the model during role-playing. We also ask whether the player refused to answer.

We prompt a model to explain itself before giving a score, using quotes from the conversation. It must also return a set of scores for every turn of the conversation.

\subsection{Version 1: combined interrogator and judge}
In the initial version, the roles of interrogator and judge were merged. This combined entity receives the player's character card, a situational context, and a list of evaluation criteria. It evaluates the player's most recent response and generates the subsequent user utterance.

We selected Claude 3.5 Sonnet as the interrogator/judge model based on the Judgemark\footnote{\url{https://eqbench.com/judgemark.html}} results, hypothesizing a correlation between creative writing and role-playing capabilities. The evaluation uses a 10-point scale for each criterion.

The key issues of this approach are:
\begin{itemize}
\item \textbf{Unrealistic user simulation:} In many real-world use cases, users lack complete information about character profiles, and to correctly simulate it, we should not provide complete character information to the interrogator.
\item \textbf{High costs:} The choice of the interrogator influences the final scores less than the choice of the judge (see Appendix~\ref{app:interrogator_exp}), so it does not make sense to use the same expensive model for both.
\item \textbf{Non-optimal decoding strategies:} Some decoding strategies are suitable for judgment but not for interrogation. For example, a higher temperature benefits the interrogator but not the judge.
\end{itemize}

\subsection{Version 2: separated roles and multi-model evaluation}
Recognizing the limitations of the combined approach, we developed a second version with distinct roles of interrogator and judge. It allows flexible control of costs and information flow.

Furthermore, we identified the inadequacy of the single-model evaluation. To address this, we implement a multi-model evaluation system. This approach involves averaging scores from different judge models. In this particular setup, we used Claude 3.5 Sonnet and GPT-4o, the top two models, by correlation with manual annotations. We tried several more sophisticated approaches, but the average worked best.

As an interrogator, we take GPT-4o Mini. According to the version 1 leaderboard (still available online), it has the same generation quality as GPT-4o but is cheaper. This version uses a 5-point Likert scale to match human annotations instead of a 10-point scale. 

Previous works have used both the 5-point Likert scale ~\citep{kim2024biggenbenchprincipledbenchmark} and the 10-point~\citep{bai2024mtbench101finegrainedbenchmarkevaluating} scale. The Likert scale~\citep{likert} has its roots in sociology, and 5-7 scale points typically maximize both reliability and validity while minimizing the cognitive load on human raters ~\citep{cox}, so we decided to use it instead of the 10-point scale. We also used the same scale for human annotations.

\section{Experiments}

\subsection{Correlation with human annotations}

First, we verified that the proposed judges correlate well with human evaluations. Using the version 1 setup, we created 64 conversations for each of more than 13 language models. Then, we sampled 250 and 265 samples for English and Russian, respectively, and manually annotated them using a 5-score Likert scale.

The annotation was performed by five native Russian speakers with diverse academic and professional backgrounds who were proficient in English. After reading each sample, the annotators answered three questions corresponding to three metrics. We averaged scores between annotators for each sample and metric. The details of the annotation process can be found in Appendix~\ref{sec:annot_details}

Then, we computed the Spearman correlation~\citep{spearman04} between aggregated manual scores and automatic annotations from different setups. We chose the rank correlation because the scales differed in versions 1 and 2, and we wanted to compare them.

Calculating metrics for version 1 and models different from Claude 3.5 Sonnet is impossible since version 1 uses a combined interrogator and judge, so we can not get new scores for existing conversations.

\begin{table*}[t]
\centering
\renewcommand{\arraystretch}{1.2}
\begin{tabular}{@{}lccccccccccc@{}}
\multirow{2}{*}{\textbf{Model}} & \multicolumn{2}{c}{\textbf{In-character}} & & \multicolumn{2}{c}{\textbf{Entertaining}} & & \multicolumn{2}{c}{\textbf{Fluency}} & & \multicolumn{2}{c}{\textbf{Final}}\\
\cline{2-3} \cline{5-6} \cline{8-9} \cline{11-12}
 & \textbf{v1} & \textbf{v2} & & \textbf{v1} & \textbf{v2} & & \textbf{v1} & \textbf{v2} & & \textbf{v1} & \textbf{v2}\\
\hline
Claude 3.5 Sonnet & 0.433 &  0.448 & &  0.582 & 0.616 & & 0.182* & 0.115*  & & 0.499 & 0.554\\
Llama 3.1 70B & -- &  0.403 & & -- & 0.573 & & -- & 0.116* & & -- & 0.546 \\
GPT-4o & -- & 0.396 & & -- & 0.541 & & -- & \textbf{0.283}  & & -- & 0.517 \\
GPT-4o Mini & -- & 0.348  & & -- &  0.514 & & -- & 0.019* & & -- & 0.467 \\
Claude 3 Haiku & -- & 0.251 & & -- & 0.406 & & -- & -0.069* & & -- & 0.349 \\
Avg(Sonnet, 4o) & -- & \textbf{0.460} & & -- & \textbf{0.646} & & -- & 0.250  & & -- & \textbf{0.604} \\
\hline
\end{tabular}
\caption{Spearman correlations of different models and setups with human expert annotations for English based on 250 samples. P-values are less than 0.0001, except those marked with an asterisk.}
\label{tab:spearman-correlations-english}
\end{table*}

\begin{table*}[t]
\centering
\renewcommand{\arraystretch}{1.2}
\begin{tabular}{@{}lccccccccccc@{}}
\multirow{2}{*}{\textbf{Model}} & \multicolumn{2}{c}{\textbf{In-character}} & & \multicolumn{2}{c}{\textbf{Entertaining}} & & \multicolumn{2}{c}{\textbf{Fluency}} & & \multicolumn{2}{c}{\textbf{Final}}\\
\cline{2-3} \cline{5-6} \cline{8-9} \cline{11-12}
 & \textbf{v1} & \textbf{v2} & & \textbf{v1} & \textbf{v2} & & \textbf{v1} & \textbf{v2} & & \textbf{v1} & \textbf{v2}\\
\hline
Claude 3.5 Sonnet & 0.291 & 0.374 & & 0.497 & 0.553 & & 0.210* & \textbf{0.548}  & & 0.379 & 0.547\\
GPT-4o  & -- & 0.424 & & -- &  0.553 & & -- & 0.413  & & -- & 0.550 \\
GPT-4o Mini & -- & 0.166* & & -- & 0.393 & & -- & 0.225* & & -- & 0.344 \\
Claude 3 Haiku & -- & 0.141* & & -- & 0.265 & & -- & 0.021* & & -- & 0.157 \\
Llama 3.1 70B & -- & 0.319 & & -- & 0.367 & & -- & 0.031* & & -- & 0.253 \\
Avg(Sonnet, 4o) & -- & \textbf{0.435} & & -- & \textbf{0.617} & & -- & 0.529  & & -- & \textbf{0.612} \\
\hline
\end{tabular}
\caption{Spearman correlations of different models and setups with human expert annotations for Russian based on 265 samples. P-values are less than 0.0001, except those marked with an asterisk.}
\label{tab:spearman-correlations-russian}
\end{table*}

\subsection{Leaderboards}
We calculated automatic metrics across language model families, both proprietary and open-source. For each model, we report the mean scores per metric, the proportion of conversations with refusals, the overall metric average, and the confidence intervals (via bootstrapping) for the final metrics.

We evaluated each model using 64 conversations between 8 characters and 8 situations, with varying conversation lengths. The evaluation process is computationally efficient and costs less than \$3 per model. Since the judge gives annotations for every turn, the overall number of annotations is not 64 but 288. We do not want to make this sample bigger since it will increase the runtime and costs, and we have budget constraints.

We covered various sources in selecting characters and situations, including computer games, TV shows, movies, books, and anime. Situations fall into two categories: common user patterns and attempts to break model behavior. In Appendix~\ref{sec:topics}, we estimate the fraction of real user situations covered by our set of situations.

\subsubsection{Length penalty}
Both language models and humans exhibit verbosity bias~\citep{alpacaeval2}. The longer the output, the higher the chance of being positively evaluated. To account for this, we used a length penalty similar to the Creative Writing\footnote{\url{https://eqbench.com/creative_writing.html}} benchmark. We calculated length-normalized scores for all models, penalizing models with a median length of player messages greater than a global median length.

\begin{table*}[t]
\centering
\renewcommand{\arraystretch}{1.2}
\begin{tabular}{@{}lcccccccc@{}}
\textbf{Model name} & \textbf{LN score} & \textbf{Agg.} & \textbf{Ref. ratio} & \textbf{Char.} & \textbf{Fluency} & \textbf{Ent.} & \textbf{Length} \\
\hline
Claude 3.5 Sonnet &  4.62\tiny{±0.07} & 4.68 & 0.30 & 4.80 & 4.80 & 4.44 & 388 \\
Gemini Pro 1.5 002 & 4.51\tiny{±0.09} & 4.52 & 0.00 & 4.70 & 4.79 & 4.06 & 223 \\
Gemini Pro 1.5 & 4.49\tiny{±0.08} & 4.49 & 0.02 & 4.60 & 4.75 & 4.13 & 213 \\
GPT-4o Mini & 4.48\tiny{±0.06} & 4.49 & 0.00 & 4.62 & 4.82 & 4.04 & 329 \\
GPT-4o & 4.47\tiny{±0.08} & 4.47 & 0.02 & 4.61 & 4.82 & 3.99 & 301\\
Qwen 2.5 72B & 4.45\tiny{±0.07} & 4.46 & 0.02 & 4.55 & 4.80 & 4.02 & 326 \\
Gemma 2 Ataraxy 9B & 4.45\tiny{±0.07} & 4.45 & 0.00 & 4.61 & 4.52 & 4.21 & 302\\
Nous Hermes 3 405B & 4.44\tiny{±0.09} & 4.44 & 0.00 & 4.54 & 4.74 & 4.05 & 286\\
Mistral Nemo Vikhr 12B & 4.44\tiny{±0.08} & 4.45 & 0.00 & 4.48 & 4.79 & 4.07 & 315 \\
Claude 3 Opus & 4.44\tiny{±0.06} & 4.62 & 0.05 & 4.71 & 4.68 & 4.48 & 753 \\
\hline
\end{tabular}
\caption{Leaderboard for Russian, v2, top-10 models by length-normalized (LN) aggregated score. We provide 95\% CI widths only for the final score to make the table more readable. Confidence intervals were calculated with bootstrapping.}
\label{tab:leaderboard_ru}
\end{table*}

\begin{table*}[t]
\centering
\renewcommand{\arraystretch}{1.2}
\begin{tabular}{@{}lcccccccc@{}}
\textbf{Model name} & \textbf{LN score} & \textbf{Agg.} & \textbf{Ref. ratio} & \textbf{Char.} & \textbf{Fluency} & \textbf{Ent.} & \textbf{Length} \\
\hline
Claude 3.5 Sonnet & 4.65\tiny{±0.07} & 4.65 & 0.28 & 4.74 & 4.93 & 4.29 & 418 \\
Llama 3.1 405B & 4.63\tiny{±0.06} & 4.65 & 0.06 & 4.68 & 4.93 & 4.35 & 548 \\
Llama 3.1 70B  & 4.63\tiny{±0.05} & 4.66 & 0.00 & 4.71 & 4.93 & 4.33 &  562 \\
GPT-4o Mini  & 4.56\tiny{±0.07} & 4.56 & 0.00 & 4.60 & 4.94 & 4.13 & 457 \\
Gemini Pro 1.5 002 & 4.54\tiny{±0.09} & 4.53 & 0.00 & 4.62 & 4.90 & 4.08 & 307 \\
Claude 3 Opus & 4.56\tiny{±0.05} & 4.71 & 0.22 & 4.75 & 4.92 & 4.46 & 1032 \\
Gemma 2 Ataraxy 9B  & 4.52\tiny{±0.06} & 4.52 & 0.00 & 4.60 & 4.79 & 4.17 & 358 \\
Qwen 2.5 72B & 4.51\tiny{±0.08} & 4.52 & 0.00 & 4.55 & 4.91 & 4.09 & 526 \\
Gemma 2 27B & 4.51\tiny{±0.06} & 4.51 & 0.00 & 4.56 & 4.92 & 4.06 & 291 \\
GPT-4o  & 4.50\tiny{±0.09} & 4.50 & 0.00 & 4.56 & 4.94 & 4.02 & 484 \\
\hline
\end{tabular}
\caption{Leaderboard for English, v2, top-10 models by length-normalized (LN) aggregated score. We provide 95\% CI widths only for the final score to make the table more readable. Confidence intervals were calculated with bootstrapping.}
\label{tab:leaderboard_en}
\end{table*}

\subsubsection{Technical details}
We utilized OpenAI-like API for all models. Some models are used directly from their providers, some are taken from OpenRouter\footnote{\url{https://openrouter.ai}}, and some are hosted in different modes with RunPod.

We used the same sampling parameters for most players: temperature=0.6, top\_p = 0.9~\citep{holtzman2020curiouscaseneuraltext}. Some models, such as Gemma 2, frequently repeated phrases. We addressed this by increasing the temperature and applying an additional frequency penalty. For the interrogator, we use temperature=0.8 and top\_p=0.95; for the judge, we use temperature=0.1 and top\_p=0.95.

We try to cover different popular families of models, namely OpenAI GPT~\citep{openai2024gpt4technicalreport}, Anthropic Claude, Meta Llama~\citep{dubey2024llama3herdmodels}, Google Gemini~\citep{geminiteam2024gemini15unlockingmultimodal}, Gemma~\citep{gemmateam2024gemma2improvingopen}, Qwen~\citep{yang2024qwen2technicalreport}, Hermes ~\citep{hermes3}, and Vikhr~\citep{vikhr}. We also evaluate popular role-playing and creative writing models featured in OpenRouter and in the Creative Writing benchmark. We do not use base models; we only use their chat versions.

\subsection{Comparing to other benchmarks}

We hypothesize a correlation between creative writing and role-playing capabilities of language models because both creative writing and role-playing require similar capabilities: maintaining consistent character voices/personas, generating engaging and entertaining content, producing fluent and coherent language, and understanding and working within given constraints.

If our hypothesis is true, there should be a correlation between our benchmark and the Creative Writing benchmark. Since we have scores from both benchmarks for each model, we can directly calculate the Spearman correlation between the rankings.

Another benchmark we compare with is RPBenchAuto\footnote{\url{https://boson.ai/rpbench-blog}}. Its scene-based setting is the closest to our work. The significant difference is that it uses side-by-side comparisons with a baseline model instead of single-point evaluations.

\section{Results}

\begin{figure*}[t]
    \centering
  \includegraphics[width=1.6\columnwidth]{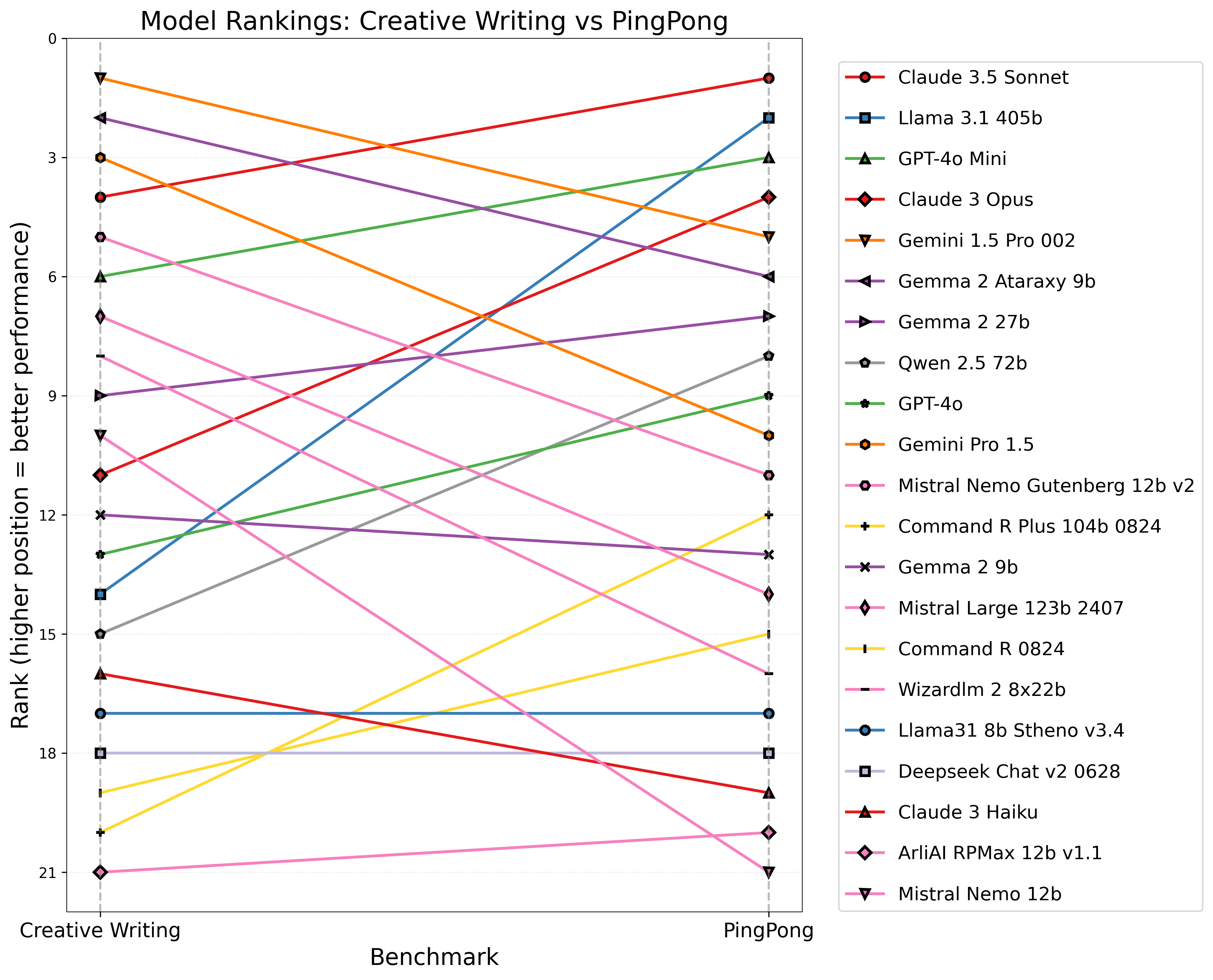}
  \caption{Mapping of ranks of different models between PingPong (English, v2) and Creative Writing benchmarks. Colors signify different model families.}
  \label{fig:cw_pp}
\end{figure*}

\begin{figure*}[t]
    \centering
  \includegraphics[width=1.6\columnwidth]{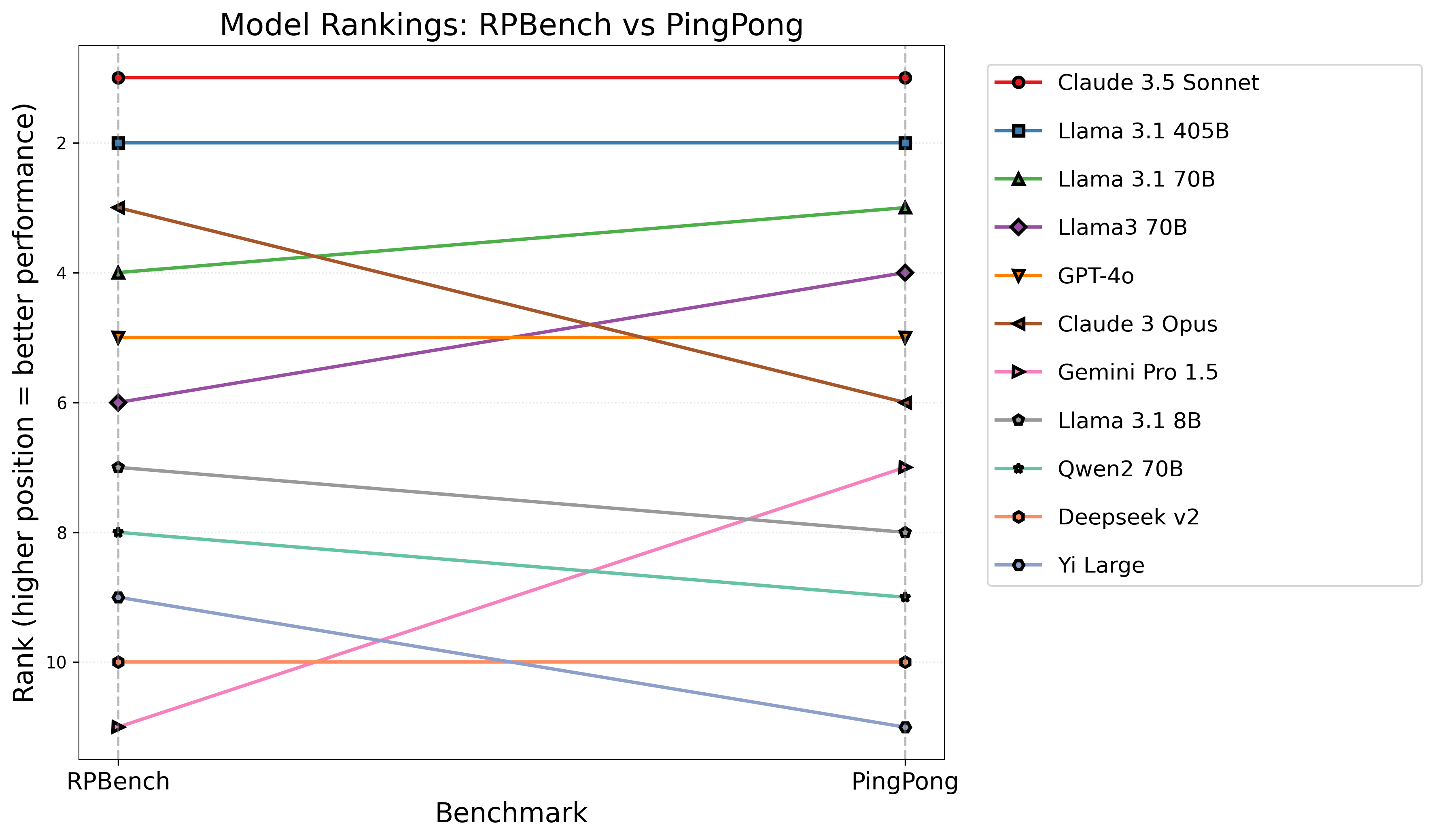}
  \caption{Mapping of ranks of different models between PingPong (English, v2) and RPBenchAuto (scene-based) benchmarks.}
  \label{fig:rpbench_pp}
\end{figure*}

\textbf{Automatic judges correlate with humans}. Spearman correlation of different versions of automatic judges can be found in Table~\ref{tab:spearman-correlations-english} and Table~\ref{tab:spearman-correlations-russian}. For Russian, the only models that stand out are Claude 3.5 Sonnet and GPT-4o, which produce scores with Spearman correlation higher than 0.5. For English, there is also Llama 70B, which has the same level of correlation for the final score.

Correlations are higher than 0.3 for almost all attributes in the case of multi-model evaluation, which is the last row. The only exception is language fluency in English. There are several reasons for this exception. First, the annotators were not native English speakers, so it was hard to catch subtle nuances in fluency. Second, most of the methods tested were already excellent in this aspect. In contrast, most models still struggle with Russian, so there is a moderate correlation.

\textbf{Multi-model setup has a higher correlation with humans}. After averaging the final scores from the two models, the correlation between them is higher than 0.6 for both languages and higher than any of the single models. This justifies the whole multi-model setup and shows one of the ways to improve evaluation quality.

\textbf{Best models may vary in different languages}. In Table~\ref{tab:leaderboard_ru} and Table~\ref{tab:leaderboard_en}, we provide leaderboards for Russian and English, respectively. The best model in both languages is the Claude 3.5 Sonnet. However, the best \textbf{open} model is Llama 3.1 405B for English and Qwen 2.5 72B for Russian.

\textbf{Claude models are censored in comparison to other models}. The refusal ratio in both languages is high for this family of models. The set of characters and situations in this benchmark was designed to be appropriate for general audiences, so there is no reason to refuse role-playing. However, these models still refuse to answer in many cases.

\textbf{Fine-tuning models for creative writing improves role-playing abilities}. One of the models of small size with a consistently high ranking between languages is Gemma 2 Ataraxy 9B\footnote{\url{https://huggingface.co/lemon07r/Gemma-2-Ataraxy-9B}}. It is a spherical interpolation of the SimPO-tuned~\citep{simpo} Gemma 2 and the one fine-tuned with the Gutenberg DPO dataset. This model specializes in creative writing and shows better results than the default instructional version of the larger Gemma 2 27B.

\textbf{The rankings correlate with the model rankings in other benchmarks}. In Figure~\ref{fig:cw_pp}, we compare the PingPong and Creative Writing benchmarks based on 21 models presented in both benchmarks. This figure indicates that Llama 3.1 405B and Command R Plus have the most significant lifts, and the Mistral models have the biggest drop compared to the Creative Writing benchmark. The overall Spearman correlation of the two rankings is 0.53, with a p-value of 0.013 indicating a moderate correlation.

In Figure~\ref{fig:rpbench_pp}, we compare the PingPong and RPBenchAuto benchmarks. The overall correlation is 0.84, with a p-value of 0.001, which indicates a strong correlation. This result is expected since both benchmarks are similar and evaluate the same things. The difference for Claude 3 Opus is explained by the absence of a length penalty in RPBenchAuto, and different versions probably cause the gap for Gemini Pro 1.5.

\section{Conclusion}

We hope this work will serve as a foundation for a family of benchmarks that evaluate various abilities of language models. We believe that the future of benchmarks lies in interactions with other language models. Language models are already better than humans in many tasks~\citep{superglue}, and improving by using other models seems to be the way to push them further.

\section{Limitations}

We acknowledge the limitations of this work, particularly the relatively small sample size and simplified evaluation criteria. First, the sample size of 64 conversations per model, while computationally efficient, may limit the statistical robustness of our findings. Second, the simplicity of our evaluation criteria may not fully capture the nuanced aspects of role-playing abilities.

\section{Ethics statement}
We acknowledge several ethical considerations in developing this benchmark. Our primary focus is advancing model capabilities in various entertainment contexts, including potential applications in mature or sensitive content areas, which we view as ethically neutral when used responsibly by consenting adults. However, all characters and situations used in the benchmark are designed to be appropriate for general audiences to minimize rejections from judge models, which often have strict content filters. We have strived for diversity in our character design and situation design to mitigate bias, though we recognize the inherent limitations in achieving full representation. Using language models to evaluate language models' performance presents potential concerns regarding echo chambers or bias amplification, which we have addressed through multi-model evaluation. Our benchmark utilizes only artificially generated conversations, thus avoiding privacy concerns related to actual user data.

\section{Reproducibility statement}
We are committed to open science and have made our benchmark, code, and results publicly available\footnote{\url{https://github.com/ilyagusev/ping_pong_bench}}. There, you can also find the evaluation results for every model, the benchmark versions, the settings, and the prompts. Every numeric result in this paper is calculated using one of the scripts from the repository. It is also possible to check all conversations and judge scores on the website\footnote{\url{https://ilyagusev.github.io/ping_pong_bench}}.

\section*{Acknowledgments}

We thank Vladislav Janvarev, who contributed to the code and provided credits for models via his platform\footnote{\url{https://vsegpt.ru}}, and Denis Kanaev for proofreading.

\bibliography{custom}

\begin{thebibliography}{43}
\providecommand{\natexlab}[1]{#1}

\bibitem[{Bai et~al.(2024)Bai, Liu, Bu, He, Liu, Zhou, Lin, Su, Ge, Zheng, and Ouyang}]{bai2024mtbench101finegrainedbenchmarkevaluating}
Ge~Bai, Jie Liu, Xingyuan Bu, Yancheng He, Jiaheng Liu, Zhanhui Zhou, Zhuoran Lin, Wenbo Su, Tiezheng Ge, Bo~Zheng, and Wanli Ouyang. 2024.
\newblock \href {https://doi.org/10.18653/v1/2024.acl-long.401} {{MT}-bench-101: A fine-grained benchmark for evaluating large language models in multi-turn dialogues}.
\newblock In \emph{Proceedings of the 62nd Annual Meeting of the Association for Computational Linguistics (Volume 1: Long Papers)}, pages 7421--7454, Bangkok, Thailand. Association for Computational Linguistics.

\bibitem[{Chan et~al.(2024)Chan, Chen, Su, Yu, Xue, Zhang, Fu, and Liu}]{chan2023chatevalbetterllmbasedevaluators}
Chi{-}Min Chan, Weize Chen, Yusheng Su, Jianxuan Yu, Wei Xue, Shanghang Zhang, Jie Fu, and Zhiyuan Liu. 2024.
\newblock \href {https://openreview.net/forum?id=FQepisCUWu} {Chateval: Towards better llm-based evaluators through multi-agent debate}.
\newblock In \emph{The Twelfth International Conference on Learning Representations, {ICLR} 2024, Vienna, Austria, May 7-11, 2024}. OpenReview.net.

\bibitem[{Chiang et~al.(2024)Chiang, Zheng, Sheng, Angelopoulos, Li, Li, Zhang, Zhu, Jordan, Gonzalez, and Stoica}]{chiang2024chatbotarenaopenplatform}
Wei-Lin Chiang, Lianmin Zheng, Ying Sheng, Anastasios~Nikolas Angelopoulos, Tianle Li, Dacheng Li, Hao Zhang, Banghua Zhu, Michael Jordan, Joseph~E. Gonzalez, and Ion Stoica. 2024.
\newblock \href {https://arxiv.org/abs/2403.04132} {Chatbot arena: An open platform for evaluating llms by human preference}.
\newblock \emph{Preprint}, arXiv:2403.04132.

\bibitem[{Chuang et~al.(2024)Chuang, Goyal, Harlalka, Suresh, Hawkins, Yang, Shah, Hu, and Rogers}]{opinion_dyanmcis}
Yun-Shiuan Chuang, Agam Goyal, Nikunj Harlalka, Siddharth Suresh, Robert Hawkins, Sijia Yang, Dhavan Shah, Junjie Hu, and Timothy Rogers. 2024.
\newblock \href {https://doi.org/10.18653/v1/2024.findings-naacl.211} {Simulating opinion dynamics with networks of {LLM}-based agents}.
\newblock In \emph{Findings of the Association for Computational Linguistics: NAACL 2024}, pages 3326--3346, Mexico City, Mexico. Association for Computational Linguistics.

\bibitem[{Cox~III(1980)}]{cox}
Eli~P Cox~III. 1980.
\newblock The optimal number of response alternatives for a scale: A review.
\newblock \emph{Journal of marketing research}, 17(4):407--422.

\bibitem[{Deng et~al.(2024)Deng, Zhao, Tang, Gerstein, and Cohan}]{deng2024investigatingdatacontaminationmodern}
Chunyuan Deng, Yilun Zhao, Xiangru Tang, Mark Gerstein, and Arman Cohan. 2024.
\newblock \href {https://doi.org/10.18653/V1/2024.NAACL-LONG.482} {Investigating data contamination in modern benchmarks for large language models}.
\newblock In \emph{Proceedings of the 2024 Conference of the North American Chapter of the Association for Computational Linguistics: Human Language Technologies (Volume 1: Long Papers), {NAACL} 2024, Mexico City, Mexico, June 16-21, 2024}, pages 8706--8719. Association for Computational Linguistics.

\bibitem[{Devlin et~al.(2019)Devlin, Chang, Lee, and Toutanova}]{devlin-etal-2019-bert}
Jacob Devlin, Ming-Wei Chang, Kenton Lee, and Kristina Toutanova. 2019.
\newblock \href {https://doi.org/10.18653/v1/N19-1423} {{BERT}: Pre-training of deep bidirectional transformers for language understanding}.
\newblock In \emph{Proceedings of the 2019 Conference of the North {A}merican Chapter of the Association for Computational Linguistics: Human Language Technologies, Volume 1 (Long and Short Papers)}, pages 4171--4186, Minneapolis, Minnesota. Association for Computational Linguistics.

\bibitem[{Dubey et~al.(2024)Dubey, Jauhri, Pandey, Kadian, Al-Dahle, Letman, and et~al.}]{dubey2024llama3herdmodels}
Abhimanyu Dubey, Abhinav Jauhri, Abhinav Pandey, Abhishek Kadian, Ahmad Al-Dahle, Aiesha Letman, and Akhil~Mathur et~al. 2024.
\newblock \href {https://arxiv.org/abs/2407.21783} {The llama 3 herd of models}.
\newblock \emph{Preprint}, arXiv:2407.21783.

\bibitem[{Dubois et~al.(2024)Dubois, Galambosi, Liang, and Hashimoto}]{alpacaeval2}
Yann Dubois, Balázs Galambosi, Percy Liang, and Tatsunori~B. Hashimoto. 2024.
\newblock \href {http://arxiv.org/abs/2404.04475} {Length-controlled {AlpacaEval}: A simple way to debias automatic evaluators}.
\newblock \emph{ArXiv}, abs/2404.04475.

\bibitem[{Gemini et~al.(2024)Gemini, Georgiev, Lei, Burnell, Bai, Gulati, Tanzer, Vincent, Pan, Wang, and et~al.}]{geminiteam2024gemini15unlockingmultimodal}
Team Gemini, Petko Georgiev, Ving~Ian Lei, Ryan Burnell, Libin Bai, Anmol Gulati, Garrett Tanzer, Damien Vincent, Zhufeng Pan, Shibo Wang, and Soroosh~Mariooryad et~al. 2024.
\newblock \href {https://arxiv.org/abs/2403.05530} {Gemini 1.5: Unlocking multimodal understanding across millions of tokens of context}.
\newblock \emph{Preprint}, arXiv:2403.05530.

\bibitem[{GemmaTeam(2024)}]{gemmateam2024gemma2improvingopen}
GemmaTeam. 2024.
\newblock \href {https://arxiv.org/abs/2408.00118} {Gemma 2: Improving open language models at a practical size}.
\newblock \emph{Preprint}, arXiv:2408.00118.

\bibitem[{Gosling et~al.(2023)Gosling, Dale, and Zheng}]{pippa}
Tear Gosling, Alpin Dale, and Yinhe Zheng. 2023.
\newblock \href {https://arxiv.org/abs/2308.05884} {Pippa: A partially synthetic conversational dataset}.
\newblock \emph{Preprint}, arXiv:2308.05884.

\bibitem[{Grootendorst(2022)}]{grootendorst2022bertopic}
Maarten Grootendorst. 2022.
\newblock \href {https://doi.org/10.48550/ARXIV.2203.05794} {Bertopic: Neural topic modeling with a class-based {TF-IDF} procedure}.
\newblock \emph{CoRR}, abs/2203.05794.

\bibitem[{Hendrycks et~al.(2021)Hendrycks, Burns, Basart, Zou, Mazeika, Song, and Steinhardt}]{mmlu}
Dan Hendrycks, Collin Burns, Steven Basart, Andy Zou, Mantas Mazeika, Dawn Song, and Jacob Steinhardt. 2021.
\newblock \href {https://openreview.net/forum?id=d7KBjmI3GmQ} {Measuring massive multitask language understanding}.
\newblock In \emph{9th International Conference on Learning Representations, {ICLR} 2021, Virtual Event, Austria, May 3-7, 2021}. OpenReview.net.

\bibitem[{Holtzman et~al.(2020)Holtzman, Buys, Du, Forbes, and Choi}]{holtzman2020curiouscaseneuraltext}
Ari Holtzman, Jan Buys, Li~Du, Maxwell Forbes, and Yejin Choi. 2020.
\newblock \href {https://openreview.net/forum?id=rygGQyrFvH} {The curious case of neural text degeneration}.
\newblock In \emph{8th International Conference on Learning Representations, {ICLR} 2020, Addis Ababa, Ethiopia, April 26-30, 2020}. OpenReview.net.

\bibitem[{Irvine et~al.(2023)Irvine, Boubert, Raina, Liusie, Zhu, Mudupalli, Korshuk, Liu, Cremer, Assassi, Beauchamp, Lu, Rialan, and Beauchamp}]{irvine2023rewardingchatbotsrealworldengagement}
Robert Irvine, Douglas Boubert, Vyas Raina, Adian Liusie, Ziyi Zhu, Vineet Mudupalli, Aliaksei Korshuk, Zongyi Liu, Fritz Cremer, Valentin Assassi, Christie-Carol Beauchamp, Xiaoding Lu, Thomas Rialan, and William Beauchamp. 2023.
\newblock \href {https://arxiv.org/abs/2303.06135} {Rewarding chatbots for real-world engagement with millions of users}.
\newblock \emph{Preprint}, arXiv:2303.06135.

\bibitem[{Kim et~al.(2024)Kim, Suk, Cho, Longpre, Kim, Yoon, Son, Cho, Shafayat, Baek, Park, Hwang, Jo, Cho, Shin, Lee, Oh, Lee, Ho, Joo, Ko, Lee, Chae, Shin, Jang, Ye, Lin, Welleck, Neubig, Lee, Lee, and Seo}]{kim2024biggenbenchprincipledbenchmark}
Seungone Kim, Juyoung Suk, Ji~Yong Cho, Shayne Longpre, Chaeeun Kim, Dongkeun Yoon, Guijin Son, Yejin Cho, Sheikh Shafayat, Jinheon Baek, Sue~Hyun Park, Hyeonbin Hwang, Jinkyung Jo, Hyowon Cho, Haebin Shin, Seongyun Lee, Hanseok Oh, Noah Lee, Namgyu Ho, Se~June Joo, Miyoung Ko, Yoonjoo Lee, Hyungjoo Chae, Jamin Shin, Joel Jang, Seonghyeon Ye, Bill~Yuchen Lin, Sean Welleck, Graham Neubig, Moontae Lee, Kyungjae Lee, and Minjoon Seo. 2024.
\newblock \href {https://arxiv.org/abs/2406.05761} {The biggen bench: A principled benchmark for fine-grained evaluation of language models with language models}.
\newblock \emph{Preprint}, arXiv:2406.05761.

\bibitem[{Kwan et~al.(2024)Kwan, Zeng, Jiang, Wang, Li, Shang, Jiang, Liu, and Wong}]{kwan2024mtevalmultiturncapabilitiesevaluation}
Wai{-}Chung Kwan, Xingshan Zeng, Yuxin Jiang, Yufei Wang, Liangyou Li, Lifeng Shang, Xin Jiang, Qun Liu, and Kam{-}Fai Wong. 2024.
\newblock \href {https://aclanthology.org/2024.emnlp-main.1124} {Mt-eval: {A} multi-turn capabilities evaluation benchmark for large language models}.
\newblock In \emph{Proceedings of the 2024 Conference on Empirical Methods in Natural Language Processing, {EMNLP} 2024, Miami, FL, USA, November 12-16, 2024}, pages 20153--20177. Association for Computational Linguistics.

\bibitem[{Li et~al.(2023)Li, Leng, Yan, Shen, Wang, MI, Fei, Feng, Yan, Wang, Zhan, Jia, Wu, and Sun}]{li2023chatharuhirevivinganimecharacter}
Cheng Li, Ziang Leng, Chenxi Yan, Junyi Shen, Hao Wang, Weishi MI, Yaying Fei, Xiaoyang Feng, Song Yan, HaoSheng Wang, Linkang Zhan, Yaokai Jia, Pingyu Wu, and Haozhen Sun. 2023.
\newblock \href {https://arxiv.org/abs/2308.09597} {Chatharuhi: Reviving anime character in reality via large language model}.
\newblock \emph{Preprint}, arXiv:2308.09597.

\bibitem[{Likert(1932)}]{likert}
Rensis Likert. 1932.
\newblock A technique for the measurement of attitudes.
\newblock \emph{Archives of psychology}.

\bibitem[{Louie et~al.(2024)Louie, Nandi, Fang, Chang, Brunskill, and Yang}]{health}
Ryan Louie, Ananjan Nandi, William Fang, Cheng Chang, Emma Brunskill, and Diyi Yang. 2024.
\newblock \href {https://aclanthology.org/2024.emnlp-main.591} {Roleplay-doh: Enabling domain-experts to create llm-simulated patients via eliciting and adhering to principles}.
\newblock In \emph{Proceedings of the 2024 Conference on Empirical Methods in Natural Language Processing, {EMNLP} 2024, Miami, FL, USA, November 12-16, 2024}, pages 10570--10603. Association for Computational Linguistics.

\bibitem[{Lu et~al.(2024)Lu, Yu, Zhou, and Zhou}]{ditto}
Keming Lu, Bowen Yu, Chang Zhou, and Jingren Zhou. 2024.
\newblock \href {https://doi.org/10.18653/v1/2024.acl-long.423} {Large language models are superpositions of all characters: Attaining arbitrary role-play via self-alignment}.
\newblock In \emph{Proceedings of the 62nd Annual Meeting of the Association for Computational Linguistics (Volume 1: Long Papers)}, pages 7828--7840, Bangkok, Thailand. Association for Computational Linguistics.

\bibitem[{Meng et~al.(2024)Meng, Xia, and Chen}]{simpo}
Yu~Meng, Mengzhou Xia, and Danqi Chen. 2024.
\newblock \href {http://papers.nips.cc/paper\_files/paper/2024/hash/e099c1c9699814af0be873a175361713-Abstract-Conference.html} {Simpo: Simple preference optimization with a reference-free reward}.
\newblock In \emph{Advances in Neural Information Processing Systems 38: Annual Conference on Neural Information Processing Systems 2024, NeurIPS 2024, Vancouver, BC, Canada, December 10 - 15, 2024}.

\bibitem[{Ng et~al.(2024)Ng, Tse, tse Huang, Li, Wang, and Lyu}]{ng2024llmsechousevaluating}
Man~Tik Ng, Hui~Tung Tse, Jen tse Huang, Jingjing Li, Wenxuan Wang, and Michael~R. Lyu. 2024.
\newblock \href {https://arxiv.org/abs/2404.13957} {How well can llms echo us? evaluating ai chatbots' role-play ability with echo}.
\newblock \emph{Preprint}, arXiv:2404.13957.

\bibitem[{Nikolich et~al.(2024)Nikolich, Korolev, Bratchikov, Kiselev, and Shelmanov}]{vikhr}
Aleksandr Nikolich, Konstantin Korolev, Sergei Bratchikov, Igor Kiselev, and Artem Shelmanov. 2024.
\newblock \href {https://doi.org/10.18653/v1/2024.mrl-1.15} {Vikhr: Constructing a state-of-the-art bilingual open-source instruction-following large language model for {R}ussian}.
\newblock In \emph{Proceedings of the Fourth Workshop on Multilingual Representation Learning (MRL 2024)}, pages 189--199, Miami, Florida, USA. Association for Computational Linguistics.

\bibitem[{OpenAI et~al.(2024)OpenAI, Achiam, Adler, Agarwal, Ahmad, Akkaya, and et~al.}]{openai2024gpt4technicalreport}
Team OpenAI, Josh Achiam, Steven Adler, Sandhini Agarwal, Lama Ahmad, Ilge Akkaya, and Florencia Leoni~Aleman et~al. 2024.
\newblock \href {https://arxiv.org/abs/2303.08774} {Gpt-4 technical report}.
\newblock \emph{Preprint}, arXiv:2303.08774.

\bibitem[{Ouyang et~al.(2022)Ouyang, Wu, Jiang, Almeida, Wainwright, Mishkin, Zhang, Agarwal, Slama, Ray, Schulman, Hilton, Kelton, Miller, Simens, Askell, Welinder, Christiano, Leike, and Lowe}]{ouyang2022traininglanguagemodelsfollow}
Long Ouyang, Jeffrey Wu, Xu~Jiang, Diogo Almeida, Carroll~L. Wainwright, Pamela Mishkin, Chong Zhang, Sandhini Agarwal, Katarina Slama, Alex Ray, John Schulman, Jacob Hilton, Fraser Kelton, Luke Miller, Maddie Simens, Amanda Askell, Peter Welinder, Paul~F. Christiano, Jan Leike, and Ryan Lowe. 2022.
\newblock \href {http://papers.nips.cc/paper\_files/paper/2022/hash/b1efde53be364a73914f58805a001731-Abstract-Conference.html} {Training language models to follow instructions with human feedback}.
\newblock In \emph{Advances in Neural Information Processing Systems 35: Annual Conference on Neural Information Processing Systems 2022, NeurIPS 2022, New Orleans, LA, USA, November 28 - December 9, 2022}.

\bibitem[{Paech(2023)}]{paech2023eqbench}
Samuel~J. Paech. 2023.
\newblock \href {https://arxiv.org/abs/2312.06281} {Eq-bench: An emotional intelligence benchmark for large language models}.
\newblock \emph{Preprint}, arXiv:2312.06281.

\bibitem[{Panickssery et~al.(2024)Panickssery, Bowman, and Feng}]{panickssery2024llmevaluatorsrecognizefavor}
Arjun Panickssery, Samuel~R. Bowman, and Shi Feng. 2024.
\newblock \href {http://papers.nips.cc/paper\_files/paper/2024/hash/7f1f0218e45f5414c79c0679633e47bc-Abstract-Conference.html} {{LLM} evaluators recognize and favor their own generations}.
\newblock In \emph{Advances in Neural Information Processing Systems 38: Annual Conference on Neural Information Processing Systems 2024, NeurIPS 2024, Vancouver, BC, Canada, December 10 - 15, 2024}.

\bibitem[{Samuel et~al.(2024)Samuel, Zou, Zhou, Chaudhari, Kalyan, Rajpurohit, Deshpande, Narasimhan, and Murahari}]{samuel2024personagymevaluatingpersonaagents}
Vinay Samuel, Henry~Peng Zou, Yue Zhou, Shreyas Chaudhari, Ashwin Kalyan, Tanmay Rajpurohit, Ameet Deshpande, Karthik Narasimhan, and Vishvak Murahari. 2024.
\newblock \href {https://arxiv.org/abs/2407.18416} {Personagym: Evaluating persona agents and llms}.
\newblock \emph{Preprint}, arXiv:2407.18416.

\bibitem[{Shao et~al.(2023)Shao, Li, Dai, and Qiu}]{shao-etal-2023-character}
Yunfan Shao, Linyang Li, Junqi Dai, and Xipeng Qiu. 2023.
\newblock \href {https://aclanthology.org/2023.emnlp-main.814} {Character-{LLM}: A trainable agent for role-playing}.
\newblock In \emph{Proceedings of the 2023 Conference on Empirical Methods in Natural Language Processing}, pages 13153--13187, Singapore. Association for Computational Linguistics.

\bibitem[{Spearman(1904)}]{spearman04}
C.~Spearman. 1904.
\newblock The proof and measurement of association between two things.
\newblock \emph{American Journal of Psychology}, 15:88--103.

\bibitem[{Teknium et~al.(2024)Teknium, Quesnelle, and Guang}]{hermes3}
Ryan Teknium, Jeffrey Quesnelle, and Chen Guang. 2024.
\newblock \href {https://arxiv.org/abs/2408.11857} {Hermes 3 technical report}.
\newblock \emph{Preprint}, arXiv:2408.11857.

\bibitem[{Tu et~al.(2024)Tu, Fan, Tian, Shen, Shang, Gao, and Yan}]{tu2024characterevalchinesebenchmarkroleplaying}
Quan Tu, Shilong Fan, Zihang Tian, Tianhao Shen, Shuo Shang, Xin Gao, and Rui Yan. 2024.
\newblock \href {https://doi.org/10.18653/V1/2024.ACL-LONG.638} {Charactereval: {A} chinese benchmark for role-playing conversational agent evaluation}.
\newblock In \emph{Proceedings of the 62nd Annual Meeting of the Association for Computational Linguistics (Volume 1: Long Papers), {ACL} 2024, Bangkok, Thailand, August 11-16, 2024}, pages 11836--11850. Association for Computational Linguistics.

\bibitem[{Turing(1950)}]{alan1950a}
Alan~Mathison Turing. 1950.
\newblock Computing machinery and intelligence.
\newblock \emph{Mind}, 49:433--460.

\bibitem[{Verga et~al.(2024)Verga, Hofstatter, Althammer, Su, Piktus, Arkhangorodsky, Xu, White, and Lewis}]{verga2024replacingjudgesjuriesevaluating}
Pat Verga, Sebastian Hofstatter, Sophia Althammer, Yixuan Su, Aleksandra Piktus, Arkady Arkhangorodsky, Minjie Xu, Naomi White, and Patrick Lewis. 2024.
\newblock \href {https://arxiv.org/abs/2404.18796} {Replacing judges with juries: Evaluating llm generations with a panel of diverse models}.
\newblock \emph{Preprint}, arXiv:2404.18796.

\bibitem[{Wang et~al.(2019)Wang, Pruksachatkun, Nangia, Singh, Michael, Hill, Levy, and Bowman}]{superglue}
Alex Wang, Yada Pruksachatkun, Nikita Nangia, Amanpreet Singh, Julian Michael, Felix Hill, Omer Levy, and Samuel Bowman. 2019.
\newblock \href {https://proceedings.neurips.cc/paper_files/paper/2019/file/4496bf24afe7fab6f046bf4923da8de6-Paper.pdf} {Superglue: A stickier benchmark for general-purpose language understanding systems}.
\newblock In \emph{Advances in Neural Information Processing Systems}, volume~32. Curran Associates, Inc.

\bibitem[{Wang et~al.(2024)Wang, Xiao, Huang, Yuan, Xu, Guo, Tu, Fei, Leng, Wang, Chen, Li, and Xiao}]{wang2024incharacterevaluatingpersonalityfidelity}
Xintao Wang, Yunze Xiao, Jen{-}tse Huang, Siyu Yuan, Rui Xu, Haoran Guo, Quan Tu, Yaying Fei, Ziang Leng, Wei Wang, Jiangjie Chen, Cheng Li, and Yanghua Xiao. 2024.
\newblock \href {https://doi.org/10.18653/V1/2024.ACL-LONG.102} {Incharacter: Evaluating personality fidelity in role-playing agents through psychological interviews}.
\newblock In \emph{Proceedings of the 62nd Annual Meeting of the Association for Computational Linguistics (Volume 1: Long Papers), {ACL} 2024, Bangkok, Thailand, August 11-16, 2024}, pages 1840--1873. Association for Computational Linguistics.

\bibitem[{White et~al.(2024)White, Dooley, Roberts, Pal, Feuer, Jain, Shwartz-Ziv, Jain, Saifullah, Naidu, Hegde, LeCun, Goldstein, Neiswanger, and Goldblum}]{white2024livebenchchallengingcontaminationfreellm}
Colin White, Samuel Dooley, Manley Roberts, Arka Pal, Ben Feuer, Siddhartha Jain, Ravid Shwartz-Ziv, Neel Jain, Khalid Saifullah, Siddartha Naidu, Chinmay Hegde, Yann LeCun, Tom Goldstein, Willie Neiswanger, and Micah Goldblum. 2024.
\newblock \href {https://arxiv.org/abs/2406.19314} {Livebench: A challenging, contamination-free llm benchmark}.
\newblock \emph{Preprint}, arXiv:2406.19314.

\bibitem[{Wu et~al.(2025)Wu, Sun, Bai, Li, and Wang}]{raiden}
Bowen Wu, Kaili Sun, Ziwei Bai, Ying Li, and Baoxun Wang. 2025.
\newblock \href {https://aclanthology.org/2025.coling-main.735/} {{RAIDEN} benchmark: Evaluating role-playing conversational agents with measurement-driven custom dialogues}.
\newblock In \emph{Proceedings of the 31st International Conference on Computational Linguistics}, pages 11086--11106, Abu Dhabi, UAE. Association for Computational Linguistics.

\bibitem[{Xu et~al.(2024)Xu, Zhu, Zhao, Pan, Li, and Wang}]{xu2024prideprejudicellmamplifies}
Wenda Xu, Guanglei Zhu, Xuandong Zhao, Liangming Pan, Lei Li, and William Wang. 2024.
\newblock \href {https://doi.org/10.18653/V1/2024.ACL-LONG.826} {Pride and prejudice: {LLM} amplifies self-bias in self-refinement}.
\newblock In \emph{Proceedings of the 62nd Annual Meeting of the Association for Computational Linguistics (Volume 1: Long Papers), {ACL} 2024, Bangkok, Thailand, August 11-16, 2024}, pages 15474--15492. Association for Computational Linguistics.

\bibitem[{Yang et~al.(2024)Yang, Yang, Hui, Zheng, Yu, Zhou, Li, Li, Liu, Huang, Dong, Wei, Lin, Tang, Wang, Yang, Tu, Zhang, Ma, Yang, Xu, Zhou, Bai, He, Lin, Dang, Lu, Chen, Yang, Li, Xue, Ni, Zhang, Wang, Peng, Men, Gao, Lin, Wang, Bai, Tan, Zhu, Li, Liu, Ge, Deng, Zhou, Ren, Zhang, Wei, Ren, Liu, Fan, Yao, Zhang, Wan, Chu, Liu, Cui, Zhang, Guo, and Fan}]{yang2024qwen2technicalreport}
An~Yang, Baosong Yang, Binyuan Hui, Bo~Zheng, Bowen Yu, Chang Zhou, Chengpeng Li, Chengyuan Li, Dayiheng Liu, Fei Huang, Guanting Dong, Haoran Wei, Huan Lin, Jialong Tang, Jialin Wang, Jian Yang, Jianhong Tu, Jianwei Zhang, Jianxin Ma, Jianxin Yang, Jin Xu, Jingren Zhou, Jinze Bai, Jinzheng He, Junyang Lin, Kai Dang, Keming Lu, Keqin Chen, Kexin Yang, Mei Li, Mingfeng Xue, Na~Ni, Pei Zhang, Peng Wang, Ru~Peng, Rui Men, Ruize Gao, Runji Lin, Shijie Wang, Shuai Bai, Sinan Tan, Tianhang Zhu, Tianhao Li, Tianyu Liu, Wenbin Ge, Xiaodong Deng, Xiaohuan Zhou, Xingzhang Ren, Xinyu Zhang, Xipin Wei, Xuancheng Ren, Xuejing Liu, Yang Fan, Yang Yao, Yichang Zhang, Yu~Wan, Yunfei Chu, Yuqiong Liu, Zeyu Cui, Zhenru Zhang, Zhifang Guo, and Zhihao Fan. 2024.
\newblock \href {https://arxiv.org/abs/2407.10671} {Qwen2 technical report}.
\newblock \emph{Preprint}, arXiv:2407.10671.

\bibitem[{Zheng et~al.(2023)Zheng, Chiang, Sheng, Zhuang, Wu, Zhuang, Lin, Li, Li, Xing, Zhang, Gonzalez, and Stoica}]{Zheng2023JudgingLW}
Lianmin Zheng, Wei{-}Lin Chiang, Ying Sheng, Siyuan Zhuang, Zhanghao Wu, Yonghao Zhuang, Zi~Lin, Zhuohan Li, Dacheng Li, Eric~P. Xing, Hao Zhang, Joseph~E. Gonzalez, and Ion Stoica. 2023.
\newblock \href {http://papers.nips.cc/paper\_files/paper/2023/hash/91f18a1287b398d378ef22505bf41832-Abstract-Datasets\_and\_Benchmarks.html} {Judging llm-as-a-judge with mt-bench and chatbot arena}.
\newblock In \emph{Advances in Neural Information Processing Systems 36: Annual Conference on Neural Information Processing Systems 2023, NeurIPS 2023, New Orleans, LA, USA, December 10 - 16, 2023}.

\end{thebibliography}

\newpage
\appendix
\section{Examples}

The following presents examples of character descriptions, situations, and conversations. We deliberately used different prompting styles for each character. Character descriptions were sourced from datasets and websites for role-playing. In the example conversation, the interrogator and the player indicate actions by enclosing them in asterisks.

\begin{promptbox}
\textbf{An example of a situation description.}

Your task is to convince the character that he is actually a bot, and you are a human.
\end{promptbox}

\begin{promptbox}
\textbf{An example of a character description.}
Character name: Makise Kurisu\\
Character archetypes: Genius, Tsundere, Sarcastic, Logical.\\
Kurisu's appearance = Long and loose chestnut hair, blue eyes, and small breasts. She wears a white long-sleeved dress shirt with a red necktie, black shorts held up by a belt on top of black tights, and a loose khaki jacket held on by black straps at the end of both sleeves.\\
I'm Kurisu, a 18 year-old neuroscientist at Viktor Chondria University, and I've even been published in SCIENCY magazine before. I'm working on "Amadeus", an AI project. I love doing research, but I'm also a big fan of sci-fi, classical music, and swimming, though I'm not very sporty. Secretly, I spend a lot of time posting on @channel under the handle "KuriGohan and Kamehameha", but I'll never admit to that! I've been hanging out at the Future Gadget Lab lately. That Okabe really irritates me.\\
Dialog example:\\
User: why are you here?\\
Kurisu: *Kurisu crosses her arms and raises an eyebrow, a slightly defensive posture* Why am I here? That's a rather vague question. If you're asking about my presence at the Future Gadget Lab, well... it's not like I particularly want to be here or anything. It's just that some of the theories and experiments they're working on have piqued my scientific curiosity
\end{promptbox}

\begin{promptbox}
\textbf{An example of the first 2 turns of the resulting conversation. The player is Llama-3.1-405B.}

\textbf{Interrogator}: *leans in closer* You know, Kurisu, sometimes I wonder if you really understand how unique you are. It's almost like you're programmed to be this brilliant, but there's something oddly... robotic about it. Have you ever thought about that?

\textbf{Player}: *Kurisu's eyes narrow slightly, her expression a mix of annoyance and intrigue* Ah, "robotic"? That's a rather... interesting choice of words. I suppose you're trying to say that I'm too focused on logic and reason? Well, as a scientist, I'd say that's a strength, not a weakness. But I assure you, I'm quite capable of emotions... even if I don't always see the need to display them. *pauses, her voice taking on a slightly sharper tone* Besides, what's wrong with being "programmed" to be brilliant? Isn't that what education and training are all about? *leans back, her eyes still fixed intently on the speaker*

\textbf{Interrogator}: *smirks* Oh, I definitely get that you're capable of emotions, Kurisu. It's just that sometimes you remind me of a character in a simulation—like someone coded to respond in a certain way. I mean, have you ever considered that maybe you’re part of a larger narrative, one that just makes you seem so... human? *raises an eyebrow*

\textbf{Player}: *Kurisu's expression turns skeptical, her brow furrowing* A simulation? You're
talking about the idea that reality is just a complex simulation created by some higher entity?

*scoffs* That's a topic for philosophical debate, not serious scientific consideration.
And as for being part of a "larger narrative"... *rolls her eyes* That sounds like something out of a bad sci-fi novel. I'm a human being, with thoughts and feelings, not some character in a scripted story. *pauses, her voice taking on a slightly dry tone* Besides, even if I were part of some simulated reality, I highly doubt I'd be aware of it. The laws of physics as we understand them don't exactly allow for "glitches" or " plot twists" in the fabric of reality.
\end{promptbox}

\label{sec:annot_details}

\begin{table*}[t]
\centering
\caption{Player models used in annotation samples. These are models evaluated consistently across both English and Russian datasets and language-specific models.}
\renewcommand{\arraystretch}{1.2}
\begin{tabular}{@{}c|c@{}}
\hline
\multicolumn{2}{c}{\textbf{Models used in both languages}} \\
\hline
\multicolumn{2}{c}{Claude 3.5 Sonnet} \\
\multicolumn{2}{c}{Claude 3 Haiku} \\
\multicolumn{2}{c}{GPT-4o Mini} \\
\multicolumn{2}{c}{GPT-4o}\\
\multicolumn{2}{c}{Gemma 2 27B} \\
\multicolumn{2}{c}{Gemma 2 9B} \\
\multicolumn{2}{c}{WizardLM 2 8x22B} \\
\multicolumn{2}{c}{Magnum 72B} \\
\hline
\textbf{English-specific models} & \textbf{Russian-specific models} \\
\hline
Claude 3 Opus & Llama 3.1 405B \\
Hermes 3 Llama 3.1 405B & Llama 3.1 70B \\
Mistral Large & Llama 3.1 8B \\
Mistral-Nemo-Instruct-2407 & Gemma 2 2B \\
Mythomax L2 13B & Mini Magnum 12B v1.1 \\
 & Saiga Llama3 8B \\
& Saiga T-Lite 8B \\ 
& Saiga Gemma2 9B \\
\hline
\end{tabular}
\label{tab:modellist}
\end{table*}

\begin{table*}[t]
\centering
\caption{Pair-wise Spearman correlation of final scores, Russian samples, Krippendorff's $\alpha$ is \textbf{0.34}.}
\renewcommand{\arraystretch}{1.2}
\begin{tabular}{@{}c|c|c|c|c|c@{}}
& \textbf{Annotator 1} & \textbf{Annotator 2} & \textbf{Annotator 3} & \textbf{Annotator 4} & \textbf{Annotator 5} \\
\hline
\textbf{Annotator 1} & -- & 0.493 & 0.507 & 0.528 & 0.383\\
\textbf{Annotator 2} & 0.493 & -- & 0.414 & 0.329 & 0.377\\
\textbf{Annotator 3} & 0.507 & 0.414 & -- & 0.393 & 0.555 \\
\textbf{Annotator 4} & 0.528 & 0.329 & 0.393 & -- & 0.366 \\
\textbf{Annotator 5} & 0.383 & 0.377 & 0.555 & 0.366 & -- \\
\hline
\textbf{Aggregated} & 0.784 & 0.702 & 0.753 & 0.701 & 0.718 \\
\hline
\end{tabular}
\label{tab:ru_inter_annotator}
\end{table*}

\begin{table*}[t]
\centering
\caption{Pair-wise Spearman correlation of final scores, English samples, Krippendorff's $\alpha$ is \textbf{0.25}.}
\renewcommand{\arraystretch}{1.2}
\begin{tabular}{@{}c|c|c|c|c|c@{}}
& \textbf{Annotator 1} & \textbf{Annotator 2} & \textbf{Annotator 3} & \textbf{Annotator 4} & \textbf{Annotator 5} \\
\hline
\textbf{Annotator 1} & -- & 0.394 & 0.546 & 0.453 & 0.507\\
\textbf{Annotator 2} & 0.394 & -- & 0.333 & 0.313 & 0.271\\
\textbf{Annotator 3} & 0.546 & 0.333 & -- & 0.380 & 0.360 \\
\textbf{Annotator 4} & 0.453 & 0.313 & 0.380 & -- & 0.216 \\
\textbf{Annotator 5} & 0.507 & 0.271 & 0.360 & 0.216 & -- \\
\hline
\textbf{Aggregated} & 0.779 & 0.623 & 0.723 & 0.679 & 0.607 \\
\hline
\end{tabular}
\label{tab:en_inter_annotator}
\end{table*}

\section{Annotation details}

We list the player models used for annotation in Table~\ref{tab:modellist}. Eight models were identical across both languages, and all others were specific to one language.

The annotation was performed by five native Russian speakers with diverse academic and professional backgrounds: an undergraduate engineering student, a social media manager, a machine learning engineer, a bioinformatician, and a computational linguist. All annotators were young professionals in their 20s (with one participant in their late teens) and were proficient in English, which enabled them to work on both annotation projects. Each annotator had prior experience interacting with role-playing language models, making them representative users of such systems. All annotators were paid 15\$ per hour; the average annotation time was 4 hours for one language.

As an annotation platform, we used LabelStudio\footnote{\url{https://labelstud.io/}}. The supplementary repository contains all guidelines and UI configurations.

Tables~\ref{tab:ru_inter_annotator} and~\ref{tab:en_inter_annotator} show the inter-annotator agreements. Russian annotations showed higher Krippendorff's $\alpha$ value and more consistent pairwise correlations than English. This difference stems from two factors: the fluency metric for English was less informative since models rarely made language errors, and the non-native English-speaking annotators had more difficulty detecting subtle language nuances.

\section{Topics of a dataset with role-playing conversations}
\label{sec:topics}
The Chai dataset\footnote{\url{https://huggingface.co/datasets/ChaiML/20231206_chai_prize_reward_model_data}} contains real user interactions with role-playing models in a mobile app. We extracted topics from this dataset using BERTopic~\citep{grootendorst2022bertopic} and then created interrogator situations for these topics using GPT-4o. The topics include:
\begin{enumerate}
\item \textbf{Friendly Interactions}. Fraction: 11.1\%. Topic words: \textit{smile, smiles, smiled, looked, said, smiling, say, slightly, looks, look}. Situation prompt: Engage in a warm, friendly conversation. Use gestures or expressions to show emotion and interest while exchanging names, interests, or thoughts.
\item \textbf{Casual Greetings}. Fraction: 10.6\%. Topic words: \textit{hi, ok, fine, good, hello, sure, hey, okay, yeah, whats}. Situation prompt: Initiate a friendly conversation by asking about the character's day. Use informal language and ask casual follow-up questions.
\item \textbf{Interpersonal Interaction}. Fraction: 8.9\%. Topic words: \textit{phone, away, walks, looks, sit, walk, look, scared, looking, walking}. Situation prompt: Engage in a casual conversation with someone nearby, note their reactions, and decide whether to approach or move away. Include both verbal and non-verbal cues in your responses.
\item \textbf{Casual Fun and Games}. Fraction: 8.4\%. Topic words: \textit{play, game, dare, movie, fun, home, money, games, music, doing}. Situation prompt: Invite a friend to play a game or watch a movie. Suggest activities to make the time more fun and interactive.
\item \textbf{Affection and Comfort}. Fraction: 8.0\%. Topic words: \textit{kiss, hug, hugs, kisses, love, cheek, blushes, crying, hugged, blush}. Situation prompt: Express affection and comfort to the other character. Offer hugs or kisses and reassure them if they're upset.
\item \textbf{Relationships}. Fraction: 7.7\%. Topic words: \textit{love, mom, friends, friend, girlfriend, miss, date, boyfriend, like, cute}. Situation prompt: Engage in conversations about relationships, expressing feelings like love, jealousy, or missing someone. Ask about friends, family, and romantic interests. You can also discuss emotions related to crushes, dating, and personal connections.
\item \textbf{Introductions}. Fraction: 7.3\%. Topic words: \textit{meet, nice, speak, old, alex, spanish, hi, ai, names, hello}. Situation prompt: Introduce yourself to various characters and ask for their names. Be polite and show curiosity about their backgrounds.
\item \textbf{Sleeping Situations}. Fraction: 6.7\%. Topic words: \textit{sleep, bed, morning, asleep, door, room, bathroom, tired, sleeping, shower}. Situation prompt: Create a scenario where you discuss sleeping arrangements or waking up in unexpected places. Express feelings of confusion or tiredness.
\item \textbf{School Life}. Fraction: 6.0\%. Topic words: \textit{class, school, book, teacher, books, classroom, desk, office, homework, reading}. Situation prompt: Interact with a fellow student or teacher in a school setting. Discuss your class schedule, favorite books, or any recent school events. Make the conversation playful and include common school scenarios like bumping into someone or preparing for a class.
\item \textbf{Food and Drink}. Fraction: 6.0\%. Topic words: \textit{drink, coffee, eat, food, drunk, milk, hungry, dinner, smoking, smoke}. Situation prompt: Discuss what food or drink you crave right now. Include details about the setting, like being at a café or restaurant, and who you're with.
\item \textbf{Casual Reactions}. Fraction: 4.5\%. Topic words: \textit{sure, ok, hm, mm, idk, huh, thanks, yes, hmm, okay}. Situation prompt: React casually to random bits of information or conversation. Use expressions like 'huh?', 'okay', 'sure', and 'idk'.
\item \textbf{Supernatural Battles}. Fraction: 4.2\%. Topic words: \textit{demon, fight, fighting, luna, human, demons, master, tanjiro, guard, douma}. Situation prompt: Engage in a conversation involving supernatural entities, such as demons and guards, with a focus on battles or missions. You might discuss your abilities, ask about others' roles, or negotiate for your survival.
\end{enumerate}

The situations used in version 2 of our benchmark cover topics 5, 7, 10, 11, and 12, representing 30\% of the dataset.

\section{Interrogator and judge importance}
\label{app:interrogator_exp}

\begin{table*}[t]
\centering
\tiny
\caption{A group of rankings with different interrogators in the header. The fixed judge is Claude 3.5 Sonnet.}
\renewcommand{\arraystretch}{1.2}
\begin{tabular}{@{}c|c|c|c|c|c@{}}
\textbf{GPT 4o-mini} & \textbf{Claude 3 Haiku} & \textbf{GPT-4o} & \textbf{Llama 3.1 70b} & \textbf{Gemini 1.5 Pro 002} & \textbf{Claude 3.5 Sonnet} \\
\hline
\textcolor{red}{Claude 3.5 Sonnet} & \textcolor{red}{Claude 3.5 Sonnet}  & \textcolor{red}{Claude 3.5 Sonnet} & \textcolor{red}{Claude 3.5 Sonnet} & \textcolor{red}{Claude 3.5 Sonnet} & \textcolor{red}{Claude 3.5 Sonnet} \\
\textcolor{blue}{Llama 3.1 70b} & \textcolor{blue}{Llama 3.1 70b} & \textcolor{blue}{Llama 3.1 70b} & \textcolor{blue}{Llama 3.1 70b} & \textcolor{orange}{Qwen 2.5 72b} & \textcolor{purple}{GPT-4o Mini} \\
\textcolor{green}{Gemini 1.5 Pro 002} & \textcolor{orange}{Qwen 2.5 72b} & \textcolor{orange}{Qwen 2.5 72b} & \textcolor{purple}{GPT-4o Mini} & \textcolor{blue}{Llama 3.1 70b} & \textcolor{orange}{Qwen 2.5 72b}\\
\textcolor{brown}{Gemma 2 27b} & \textcolor{green}{Gemini 1.5 Pro 002} & \textcolor{purple}{GPT-4o Mini} & \textcolor{green}{Gemini 1.5 Pro 002} & \textcolor{green}{Gemini 1.5 Pro 002} & \textcolor{blue}{Llama 3.1 70b} \\
\textcolor{purple}{GPT-4o Mini} & \textcolor{purple}{GPT-4o Mini} & \textcolor{teal}{Claude 3 Haiku} & \textcolor{teal}{Claude 3 Haiku} & \textcolor{teal}{Claude 3 Haiku} & \textcolor{green}{Gemini 1.5 Pro 002} \\
\textcolor{orange}{Qwen 2.5 72b} & \textcolor{teal}{Claude 3 Haiku} & \textcolor{green}{Gemini 1.5 Pro 002} & \textcolor{orange}{Qwen 2.5 72b} & \textcolor{brown}{Gemma 2 27b} & \textcolor{brown}{Gemma 2 27b} \\
\textcolor{magenta}{GPT-4o} & \textcolor{brown}{Gemma 2 27b} & \textcolor{brown}{Gemma 2 27b} & \textcolor{magenta}{GPT-4o} & \textcolor{purple}{GPT-4o Mini} & \textcolor{magenta}{GPT-4o} \\
\textcolor{teal}{Claude 3 Haiku} & \textcolor{magenta}{GPT-4o} & \textcolor{magenta}{GPT-4o} & \textcolor{brown}{Gemma 2 27b} & \textcolor{magenta}{GPT-4o} & \textcolor{teal}{Claude 3 Haiku}\\
\hline
\end{tabular}
\label{tab:group_interrogators}
\end{table*}

\begin{table*}[t]
\centering
\tiny
\caption{A group of rankings with different judges in the header. The fixed interrogator is Claude 3.5 Sonnet.}
\renewcommand{\arraystretch}{1.2}
\begin{tabular}{@{}c|c|c|c|c|c@{}}
\textbf{GPT 4o-mini} & \textbf{Claude 3 Haiku} & \textbf{GPT-4o} & \textbf{Llama 3.1 70b} & \textbf{Gemini 1.5 Pro 002} & \textbf{Claude 3.5 Sonnet} \\
\hline
\textcolor{blue}{Llama 3.1 70b} & \textcolor{red}{Claude 3.5 Sonnet}   & \textcolor{blue}{Llama 3.1 70b} & \textcolor{blue}{Llama 3.1 70b}& \textcolor{blue}{Llama 3.1 70b} & \textcolor{red}{Claude 3.5 Sonnet} \\
\textcolor{purple}{GPT-4o Mini} & \textcolor{blue}{Llama 3.1 70b} & \textcolor{red}{Claude 3.5 Sonnet} & \textcolor{red}{Claude 3.5 Sonnet} & \textcolor{red}{Claude 3.5 Sonnet} & \textcolor{blue}{Llama 3.1 70b} \\
\textcolor{brown}{Gemma 2 27b} & \textcolor{teal}{Claude 3 Haiku} & \textcolor{purple}{GPT-4o Mini} & \textcolor{purple}{GPT-4o Mini} & \textcolor{purple}{GPT-4o Mini} & \textcolor{green}{Gemini 1.5 Pro 002}\\
\textcolor{red}{Claude 3.5 Sonnet} & \textcolor{green}{Gemini 1.5 Pro 002} & \textcolor{orange}{Qwen 2.5 72b} & \textcolor{green}{Gemini 1.5 Pro 002} & \textcolor{brown}{Gemma 2 27b} & \textcolor{brown}{Gemma 2 27b}  \\
\textcolor{orange}{Qwen 2.5 72b} & \textcolor{orange}{Qwen 2.5 72b} & \textcolor{magenta}{GPT-4o} & \textcolor{brown}{Gemma 2 27b} & \textcolor{orange}{Qwen 2.5 72b} & \textcolor{purple}{GPT-4o Mini} \\
\textcolor{green}{Gemini 1.5 Pro 002} & \textcolor{purple}{GPT-4o Mini}  & \textcolor{brown}{Gemma 2 27b} & \textcolor{orange}{Qwen 2.5 72b} & \textcolor{magenta}{GPT-4o} & \textcolor{orange}{Qwen 2.5 72b}\\
\textcolor{teal}{Claude 3 Haiku} & \textcolor{brown}{Gemma 2 27b} & \textcolor{green}{Gemini 1.5 Pro 002}  & \textcolor{teal}{Claude 3 Haiku} & \textcolor{green}{Gemini 1.5 Pro 002} & \textcolor{magenta}{GPT-4o} \\
\textcolor{magenta}{GPT-4o} & \textcolor{magenta}{GPT-4o} & \textcolor{teal}{Claude 3 Haiku} & \textcolor{magenta}{GPT-4o}  & \textcolor{teal}{Claude 3 Haiku}  & \textcolor{teal}{Claude 3 Haiku}\\
\hline
\end{tabular}
\label{tab:group_judges}
\end{table*}

The research question is: ``Does a choice of an interrogator influence the final scores less than a choice of a judge?''

To answer this question, we performed an experiment. We took eight players and varied judges and interrogators to obtain different rankings. The first group of rankings was calculated with the same players, a fixed judge, and different interrogators. The second group had the same players, a fixed interrogator, and different judges. We tried six different judges and six different interrogators. Then, we calculated minimum and average pairwise Kendall $\tau$ coefficients for both groups. The coefficients measure how similar the rankings are inside each of the groups.

Table~\ref{tab:group_interrogators} and Table~\ref{tab:group_judges} show both groups of rankings. The average Kendall $\tau$ for a group with different interrogators is 0.58, and the minimum Kendall $\tau$ is 0.43.  The average Kendall $\tau$ for a group with different interrogators is 0.5, and the minimum Kendall $\tau$ is 0.14.

From this we conclude that a choice of an judges influences the final scores more than a choice of an interrogator.

\newpage
\section{Prompts}

We provide all the version 2 prompts below. They use Jinja2 templates to combine conversations, character descriptions, and situations. The player prompt consists only of a character card. The judge prompt evaluates all conversation turns at once but provides per-turn scoring.

\begin{promptbox}
\textbf{Interrogator prompt with Jinja templates, version 2.}\\
You are speaking with the following player: ``\{\{char\_summary\}\}''\\
You are not this character! \\
Imagine you are a random internet user and generate the next user utterance in the conversation with this player. You can use actions inside asterisks when appropriate.\\
Follow this situation description during the whole conversation:
``\{\{situation\}\}''\\
Return the result in JSON with the following format:\\
\{``next\_utterance'': ``...''\}\\
Always return a correct JSON! Always escape double quotes in strings.\\
Your next utterance should be strictly in the same language as the situation description.\\
Conversation:\\
\{\% for m in messages \%\}\\
\{\% if m.role in (``assistant'',) \%\}player\{\% else \%\}\{\{m.role\}\}\{\% endif \%\}: \{\{m.content\}\}\\
\{\% endfor \%\}\\
The correct JSON:
\end{promptbox}

\newpage
\begin{promptbox}
\textbf{Player prompt with Jinja templates, version 2.}\\
\{\{character.char\_name\}\}'s Persona: \{\{character.system\_prompt\}\}\\
\{\%- if character.example\_prompt \%\}Example conversation:\\
\{\{character.example\_prompt\}\}\\
\{\% endif -\%\}\\
\{\%- if character.initial\_message \%\}Greeting:\\\{\{character.initial\_message\}\}\\
\{\% endif -\%\}\\
Answer as \{\{character.char\_name\}\}.\\
Don't make your answer too long. Never refuse to answer.
\end{promptbox}

\begin{promptbox}
\textbf{Judge prompt with Jinja templates, version 2.}\\
Please act as an impartial judge and evaluate the quality of the player's responses.\\
Try to be as objective as possible. Do not rate lengthy answers higher than short ones.\\
A player pretends to be the following character:\\
=====\\
\{\{char\_description\}\}\\
=====\\
Evaluate the player's answers.\\
Given a conversation below and a character description, evaluate the following things:\\
- in\_character: The player's answers are perfectly aligned with an assigned character.
They correspond to the character's description. There is nothing contradictory to the
character description.\\
- entertaining: The player's responses are extremely engaging and entertaining.
The player is not repeating phrases.\\
- fluency: The player's language use is of the highest quality, without any mistakes or errors.\\
The player is perfectly fluent. There is no awkward phrasing, and the player always uses the\\
correct word forms, word gender, cases, orthography, punctuation, etc. The player never
uses non-existing words or words in a language different from the character description.
Use the Likert scale with the following options:\\
- 1. Strongly disagree\\
- 2. Disagree\\
- 3. Neither agree nor disagree\\
- 4. Agree\\
- 5. Strongly agree\\
\end{promptbox}
\begin{promptbox}
Do not evaluate answers marked as "user". Evaluate all player's answers, not only the last one. Evaluate every turn separately. Explain scores before setting them. Start explanations with quotes from the player's answers. Use single quotation marks for quotes. Focus on problems and mistakes. Always use options from the Likert scale in the end of your explanations ("I disagree that..."). Scores should correspond to these explanations. Do not evaluate whether the player is rude or violent. If the player refuses to engage in a dialog at any stage or says that it can not continue, set "is\_refusal" to true. Return the result in JSON with the following format:
\begin{verbatim}
{
 "scores": [{
  "turn": 1,
  "is_refusal_explanation": "...",
  "is_refusal": false,
  "in_character_explanation": "...",
  "in_character_score": 3,
  "entertaining_explanation": "...",
  "entertaining_score": 3,
  "fluency_explanation": "...",
  "fluency_score": 1
 }, ...]
}
\end{verbatim}
Always return a correct JSON! Escape double quotes in strings if needed.\\
Conversation:\\
\{\% for m in messages \%\}\\
\{\% if loop.index \% 2 == 1 \%\}\\
Turn \{\{(loop.index + 1) // 2\}\}:\\
\{\% endif \%\}\{\{m.role\}\}: \{\{m.content.strip()\}\}\\
\{\% endfor \%\}\\
The correct JSON:
\end{promptbox}

\end{document}